\documentclass[10pt,conference,letterpaper]{IEEEtran}

\usepackage[utf8]{inputenc}

\usepackage{graphicx}
\usepackage{listing}
\usepackage{amsopn}
\usepackage{amsmath}

\usepackage{cite}

\usepackage[absolute]{textpos}

\begin{document}

\begin{textblock}{13}(2.5,0.5)
Pre-print version. Full version of this work will be published in the proceedings of IJCNN 2018
\end{textblock}

\title{A Bimodal Learning Approach to Assist Multi-sensory Effects Synchronization}

\author{\IEEEauthorblockN{Raphael Abreu}
\IEEEauthorblockA{CEFET/RJ\\
Rio de Janeiro, Brazil\\
raphael.abreu@eic.cefet-rj.br}
\and
\IEEEauthorblockN{Joel dos Santos}
\IEEEauthorblockA{CEFET/RJ\\
Rio de Janeiro, Brazil\\
jsantos@eic.cefet-rj.br}
\and
\IEEEauthorblockN{Eduardo Bezerra}
\IEEEauthorblockA{CEFET/RJ\\
Rio de Janeiro, Brazil\\
ebezerra@cefet-rj.br}}

\maketitle

\begin{abstract}
In mulsemedia applications, traditional media content (text, image, audio, video, etc.) can be related to media objects that target other human senses (e.g., smell, haptics, taste). Such applications aim at bridging the virtual and real worlds through sensors and actuators. Actuators are responsible for the execution of sensory effects (e.g., wind, heat, light), which produce sensory stimulations on the users. In these applications sensory stimulation must happen in a timely manner regarding the other traditional media content being presented. For example, at the moment in which an explosion is presented in the audiovisual content, it may be adequate to activate actuators that produce heat and light. It is common to use some declarative multimedia authoring language to relate the timestamp in which each media object is to be presented to the execution of some sensory effect. One problem in this setting is that the synchronization of media objects and sensory effects is done manually by the author(s) of the application, a process which is time-consuming and error prone. In this paper, we present a bimodal neural network architecture to assist the synchronization task in mulsemedia applications. Our approach is based on the idea that audio and video signals can be used simultaneously to identify the timestamps in which some sensory effect should be executed. Our learning architecture combines audio and video signals for the prediction of scene components. For evaluation purposes, we construct a dataset based on Google's AudioSet. We provide experiments to validate our bimodal architecture. Our results show that the bimodal approach produces better results when compared to several variants of unimodal architectures.
\end{abstract}

\section{Introduction} 
\label{sec:introduction}

Multimedia applications involve the presentation of different audiovisual objects organized in time and space \cite{Blakowski:1996aa,Hardman:1998zj}. Given the nature of the content they present, the majority of current multimedia applications stimulate only two human senses: sight and hearing. As discussed in \cite{Ghinea:2014aa}, aiming at increasing the user quality of experience (QoE) and immersion with multimedia applications, the literature present works \cite{Waltl:2010aa,Rainer:2012aa,Yuan:2015aa} that propose the use of other sensory effects in multimedia applications in order to provide users with new sensations during a multimedia presentation. In \cite{Ghinea:2014aa}, the term mulsemedia (\emph{MULtiple SEnsorial MEDIA}) is put forward to denote multimedia applications in which traditional media content (text, image, audio, video, etc.) can be related to media objects that target other human senses (e.g., smell, haptics, etc.).

To clarify, let us describe a simple yet powerful example\footnote{The application we describe here is inspired by the Day's Route application available at http://club.ncl.org.br/node/69}. Consider a non-linear show, i.e., a show whose narrative line is not known \textit{a priori} and is constructed based on user interaction. In this show the user actively participates in the construction of the narrative line by choosing the next sight in a city tour from a list of available options provided by the application. At each sight, a video and complementary information about it are presented to the user. At the beginning of the show, the user may choose whether he/she wants to interact with the application. If not, a default tour is presented. Let us now consider an evolution of such application which includes sensory effects. In this new application, several sensory effects are presented along the narrative line in a synchronized way, with the purpose of increasing the immersion of the user in the audiovisual content. In this scenario, if the user chooses to visit a particular sight, e.g., the beach at Rio de Janeiro, the sensory effects to be presented by the application would mimic the sight's environmental conditions (e.g., hot wind blowing, smell of the sea breeze, etc).

Synchronization plays an important role for multimedia applications. Multimedia authoring languages are domain specific languages that provide constructions for defining how media objects shall be presented during the execution of a multimedia application, i.e., their temporal synchronization. Moreover, those languages also manage user interaction as a special case of temporal synchronization. Examples of such languages are 
SMIL (\emph{Synchronized Multimedia Integration Language})~\cite{W3C:2008zk} and NCL (\emph{Nested Context Language})~\cite{ITU:2009ma}.

When considering the use of sensory effects in multimedia applications, the usual approach is to use audiovisual media objects as the base for synchronization, such that the timestamps in which sensory effects are to be executed are defined in relation to certain media objects. For example, light and heat effects may be presented when an explosion occurs in the main video. One should notice, however, that although multimedia languages easy the specification of the synchronization aspect in an application, the task of defining all the moments in which a given sensory effect shall be executed is still carried manually by the author. In general, authors relate sensory effects to the content being presented by the application. 

In \cite{Abreu:2017ab}, we define a \emph{scene component} as, a given element (rock, tree, dog, person, etc.) or concept (happy, crowded, dark, etc.) that appears in the content of a media object. In the application example we presented earlier, scene components may refer to the sun, the beach, trees, flowers, and other elements that may appear in the Rio de Janeiro's sights presented in the touristic program.

In previous work \cite{Abreu:2017ab,Abreu:2017aa}, we tackle the problem of automatically recognizing scene components in audiovisual objects, in order to assist the realization of the synchronization task in mulsemedia applications. We proposed an architecture capable of identifying the presence of scene components in video and audio objects and defining, in a semi-supervised manner, the synchronization among sensory effects and an application main video and/or audio.

In this paper, we focus on the recognition of scene components specifically related to sensory effects. Examples of such kind of scene components are wind, explosion, rain, lightning, etc. Some  of these components may be predominantly found either in the video (e.g., lightning) or in the audio (e.g., wind). Other components (e.g, explosion) can be found in both modalities of the audiovisual object. In particular, we propose the use of a bimodal neural network architecture for increasing the accuracy of recognition of scene components specifically for the task of synchronizing sensory effects in mulsemedia applications. Our premise is that trying to combine both modalities in the identification can produce a better accuracy when compared to the separate identification. We perform computational experiments with different network architectures, both unimodal (either audio or video) and bimodal (audio and video). Through experimental verification, we show that the proposed bimodal fusion approach is superior in accuracy when compared with any single modal recognition system.

The contributions of this paper are twofold: (i) we propose a bimodal learning architecture for predicting scene components in audiovisual content, and (ii) we present the first version of a dataset tailored for the prediction of scene components related to sensory effects.

The remaining of this paper is structured as follows.
Section~\ref{sec:related_work} presents related work regarding the use of bimodal neural networks to combine audio and video information in several application domains.
Section~\ref{sec:dataset} provides a brief description of AudioSet, the collection of video clips we use on our learning architecture. This Section also describes how we built our own training set based on AudioSet.
Section~\ref{sec:network_architectures} discusses the network architectures used in this work and how both modalities can be used to improve accuracy.
Section~\ref{sec:experiments_and_results} presents experimental results along with a corresponding analysis.
Section~\ref{sec:conclusion} concludes and discusses future work.

\section{Related Work} 
\label{sec:related_work}

Over the last few years, several neural network architectures have been proposed as a fusion model for audio and visual data, in a plethora of application domains, such as speech recognition~\cite{NgiamKKNLN11,DBLP:journals/access/TorfiIND17,7965918}, speaker recognition~\cite{7814493,Hu:2015:DMS:2733373.2806293,5280184}, video classification~\cite{Yang:2016:MMF:2964284.2964297}, emotion recognition~\cite{Zhang:2016:MDC:2911996.2912051}, natural sound recognition~\cite{NIPS2016:6146,BODDAPATI20172048}, and person recognition~\cite{Vegad:10.4172/2155-6180.1000377}.

One of the first attempts of using neural networks to explore the correlation between audio and visual information was made by Ngiam et al~\cite{NgiamKKNLN11}. They use an extension of Restricted Boltzmann Machines with sparsity to learn better single modality representations given unlabeled data from multiple modalities.  They apply their model to improve the accuracy of speech recognition by using not only the audio signal, but also the image frames of lips of the speaker. The authors show that better features for one modality (e.g., video) can be learned if multiple modalities (e.g., audio and video) are used at feature learning time. Also aiming at leveraging images of people speaking, in \cite{DBLP:journals/access/TorfiIND17} the authors use 3D convolutional neural networks for audio-visual matching in which a bridge between spatio-temporal features is established to build a common feature space between audio-visual modalities. Another architecture to learn a model for audio visual speech recognition is proposed in \cite{7965918}. This model comprises a convolutional neural network (CNN) followed by a Long Short-Term Memory (LSTM) neural network to handle visual modality, another LSTM RNN to handle audio modality, and a multimodal layer to fuse the outputs of both modalities.

In \cite{7814493}, an audio-visual speaker recognition method is presented. The method works by fusing face and audio information via multi-modal correlated neural networks. The  facial  features are learned  by  convolutional  neural networks. In \cite{Hu:2015:DMS:2733373.2806293}, the authors use a CNN as a face feature extractor from face imagery data, which are latter stacked with mel frequency cepstrum coefficients. In \cite{5280184}, the authors present a model to learn bimodally informative structures from audio–visual signals. They approach the problem of multimodal data processing by representing each signal as a sparse sum of audio–visual kernels. The authors also propose an unsupervised learning algorithm to form dictionaries of bimodal kernels from audio–visual material.

In \cite{Yang:2016:MMF:2964284.2964297} a multilayer and multimodal fusion framework of deep neural networks for video classification is proposed. The authors use four modalities to extract complementary information across multiple temporal scales. For each single modality, discriminative representations are computed for convolutional and fully connected layers. For the fusion of multiple layers and modalities, they propose an adaptive boosting model to learn the optimal combination of them. In contrast to our current work, the authors do not use audio modality.

In \cite{Zhang:2016:MDC:2911996.2912051}, the authors propose a multimodal Deep Convolution Neural Network (DCNN) to combine audio and visual cues for emotion recognition. They first fine-tune two DCNN models to perform audio and visual emotion recognition tasks respectively on the corresponding labeled speech and face data. Then, the outputs of these two DCNNs are integrated in a fully-connected neural network, which is trained to obtain a joint audio-visual feature representation for emotion recognition.

In \cite{NIPS2016:6146}, SoundNet is presented, a deep convolutional architecture  for natural sound recognition. SoundNet learns audio representations directly on raw audio signals. The audio recognition model is trained by transferring knowledge from pretrained visual representations and large amounts of unlabelled video. The authors achieve state-of-the-art accuracy on three standard acoustic scene classification datasets. Another approach to recognize natural (environmental) sound is presented in \cite{BODDAPATI20172048}, in which the author's convolutional neural networks are designed specifically for object recognition in images, and can be successfully trained to classify spectral images of environmental sounds. Similar to our current work, both \cite{NIPS2016:6146} and \cite{BODDAPATI20172048} try to classify non-human sounds. However, they use only the sound modality, not exploiting information from other modalities to help identifying the sound. Moreover, they do not aim at assisting synchronization task in mulsemedia applications.

In \cite{Vegad:10.4172/2155-6180.1000377}, the authors propose an audio-visual bimodal person recognition system. This system uses CNNs as a primary model architecture. First, two separate Deep CNN models are trained with the help of audio and facial features, respectively. The outputs of these CNN models are then combined/fused to predict the identity of the subject (person).



\section{Dataset Construction}
\label{sec:dataset}

AudioSet~\cite{45857} is an extensive collection of 10-second segments (clips) of sound that belongs to YouTube videos. Each segment comes annotated with audio events found in it. This dataset contains $632$ audio event classes and over 2 million sound clips. The dataset is divided in three disjoint sets: a balanced training set, an evaluation set and a unbalanced training set. The first two sets are balanced so that every label is associated to roughly $60$ examples. The unbalanced training set contains the remainder examples in the collection. AudioSet does not actually come with the video contents. Instead, each entry in this catalog makes a reference to the corresponding video ID, whose content can in turn be accessed from YouTube.

Although AudioSet does not provide video features, it provides a CSV file that ties the YouTube video ID to the audio events found in the segment. For example, the entry \textsc{7Zdx0YrzHVk,20.000,30.000,``/m/02\_41,/m/0838f''} in that CSV file means the YouTube video whose ID is \textsc{7Zdx0YrzHVk}, for the 10 seconds timeframe from $20s$ to $30s$, presents sounds of ``Fire'' (\textsc{/m/02\_41}) and ``Water'' (\textsc{/m/0838f}). With this information, we could select the related videos and download them (from YouTube), for every label in our chosen subset of AudioSet. In the remaining of this section, we describe how we constructed our training and validation datasets taking AudioSet as starting point.

The training dataset we constructed for use in our experiments (see Section~\ref{sec:experiments_and_results}) is a subset of the clips referenced by AudioSet. In particular, we used the unbalanced training set already provided by AudioSet for building our own training set. This subset was built by applying a two-step selection procedure, as described in the following paragraphs. 

\begin{figure}[htb]
    \includegraphics[width=0.45\textwidth]{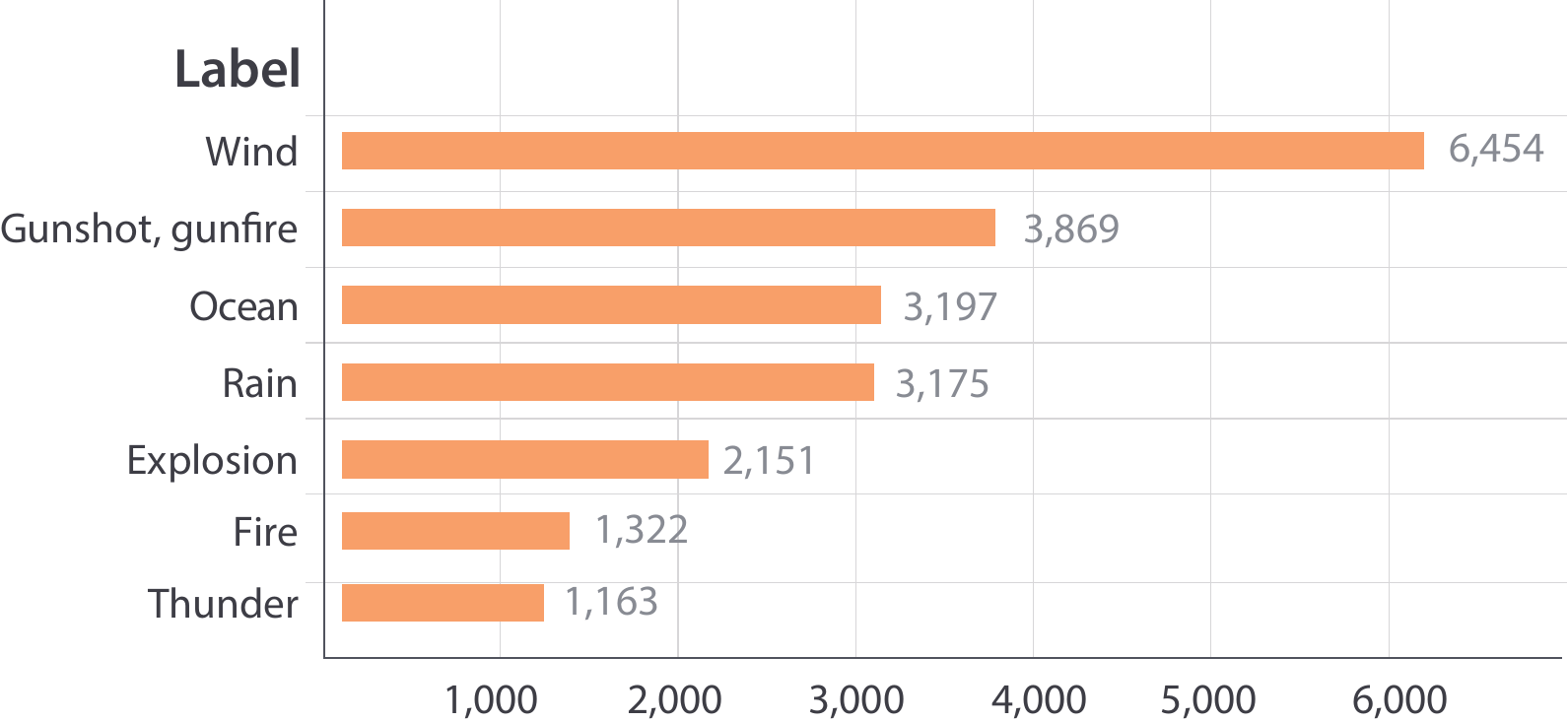}
    \caption{Distribution of labels in the unbalanced training dataset of AudioSet for the chosen labels (\textit{Wind}, \textit{Thunder}, \textit{Rain}, \textit{Ocean}, \textit{Fire}, \textit{Explosion} and \textit{Gunshot, gunfire}).}
    \label{fig:histograma_audioset}
\end{figure}

In the first step, we selected the examples associated to a small subset of the $632$ audio event classes contained in AudioSet. Since in this work we are interested in sound events that can be associated to sensory effects, we selected only examples coming from the following 7 labels for event classes from AudioSet: \textit{Wind}, \textit{Thunder}, \textit{Rain}, \textit{Ocean}, \textit{Fire}, \textit{Explosion} and \textit{Gunshot, gunfire}.  The overall distribution of these 7 labels in the original dataset is shown in Fig.~\ref{fig:histograma_audioset}. Note that the distribution is not uniform as this dataset is unbalanced.

\begin{figure}[htb]
    \includegraphics[width=0.45\textwidth]{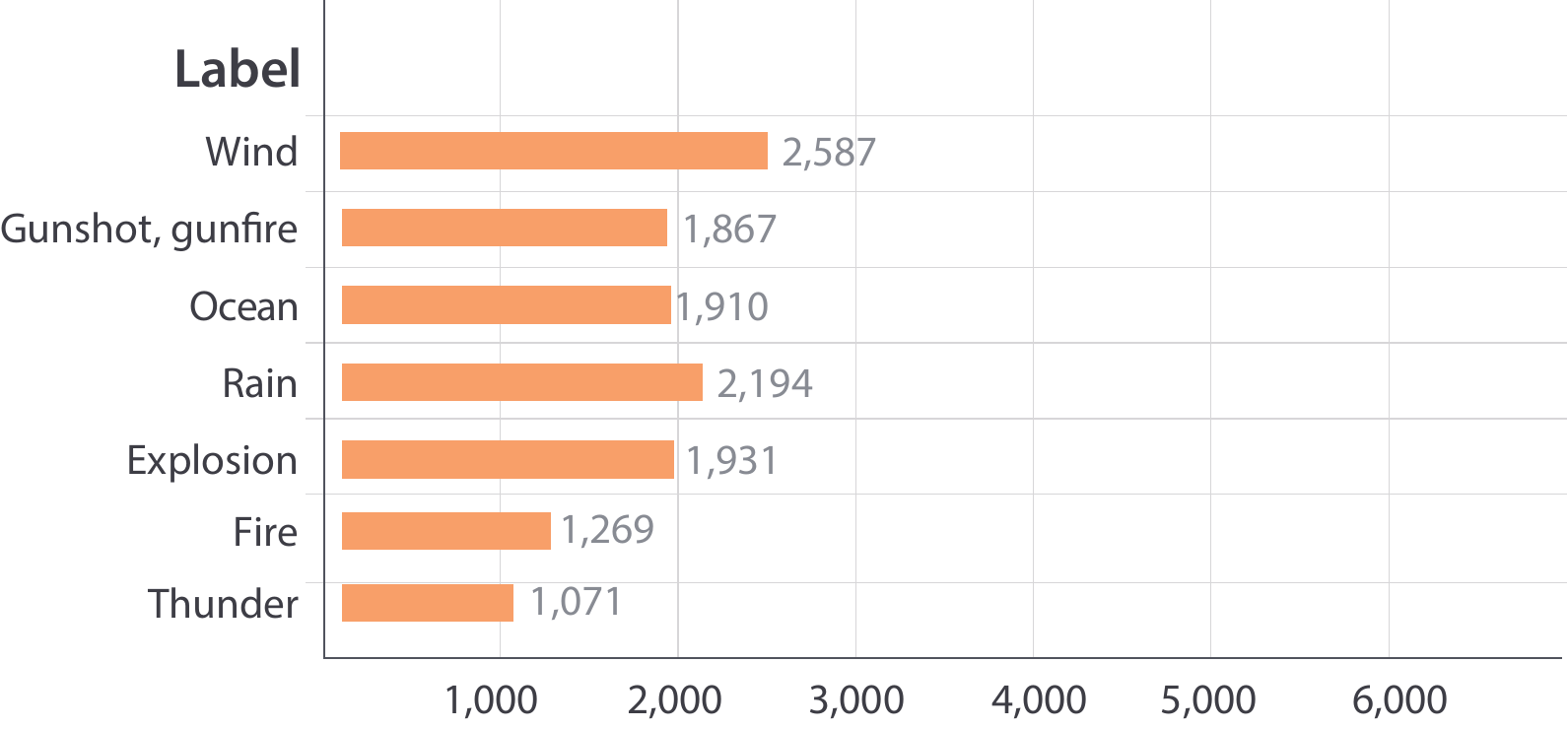}
    \caption{Distribution of segments in our training dataset, for the chosen labels: \textit{Wind} ($20.17\%$), \textit{Thunder} ($8.35\%$), \textit{Rain} ($17.1\%$), \textit{Ocean} ($14.89\%$), \textit{Fire} ($9.89\%$), \textit{Explosion} ($15.05\%$) and \textit{Gunshot, gunfire} ($14.55\%$).}
    \label{fig:histograma_subset}
\end{figure}

In the second step, we took as input the examples resulting from the first step. In order to alleviate the existing imbalance in the AudioSet subset associated to the above selected labels, we fixed an upper limit to download at most $2000$ examples for each of the 7 labels that we used. However, due to label co-occurrence in the data, the actual number of downloaded examples for some labels (namely, \emph{Rain} and \emph{Wind}) ended up to be higher than the upper limit we established. In total, we downloaded $11,518$ distinct segments. A total of $568$ videos failed to download (which accounts for a degradation of $\approx~5\%$). The cause for this failure may be that the videos were not accessible from our location, or that these videos have been removed. Thus the resulting size of our training set is $10,950$ distinct segments with a total of $12,829$ labels. Fig.~\ref{fig:histograma_subset} shows the resulting distribution of retrieved segments for our training dataset.

\begin{figure}[htb]
    \includegraphics[width=0.45\textwidth]{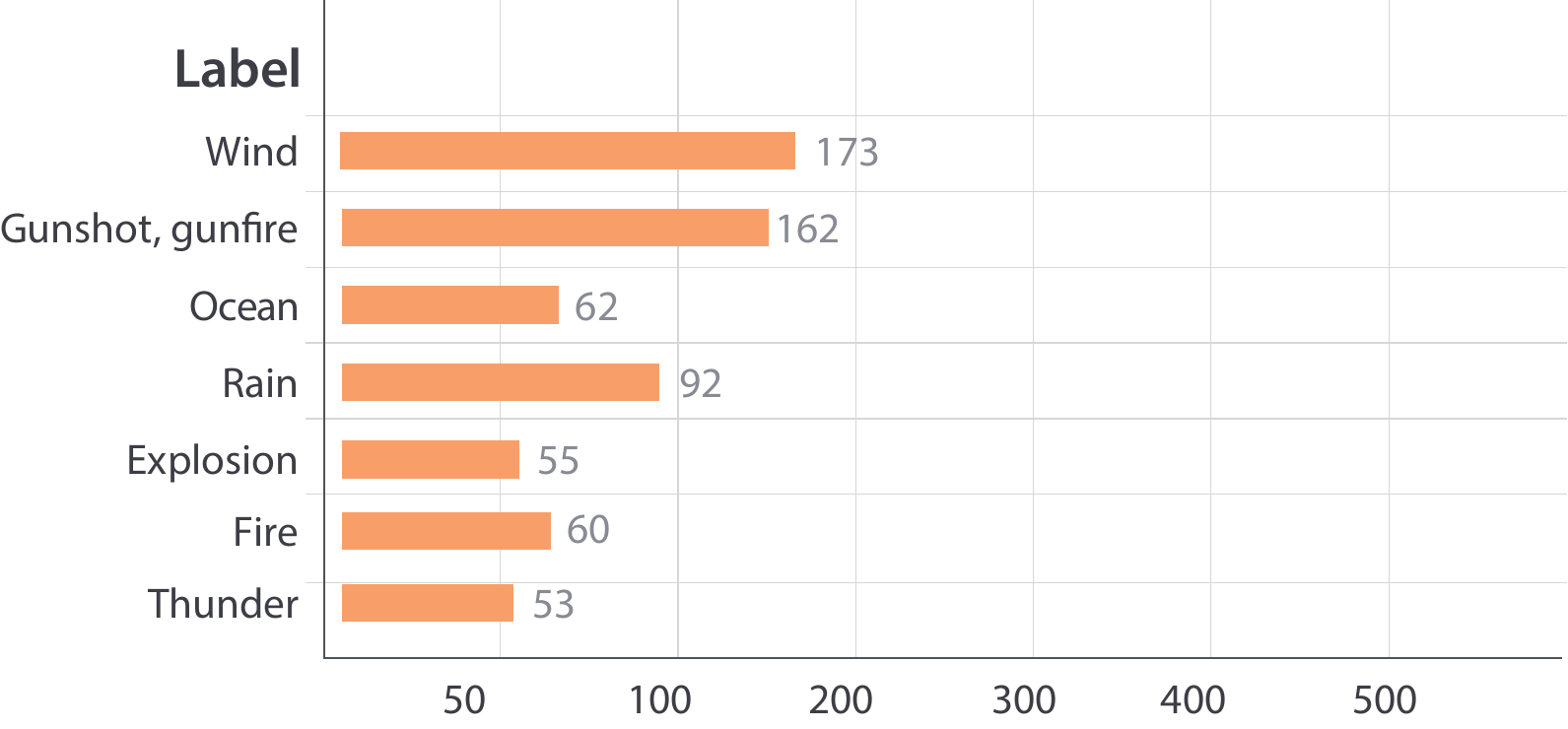}
    \caption{Distribution of segments in our validation dataset, for the chosen labels: \textit{Wind} ($26.33\%$), \textit{Thunder} ($8.07\%$), \textit{Rain} ($14.0\%$), \textit{Ocean} ($9.44\%$), \textit{Fire} ($9.13\%$), \textit{Explosion} ($8.37\%$) and \textit{Gunshot, gunfire} ($24.66\%$).
    }
    \label{fig:histograma_audioset_validation}
\end{figure}

In order to build our validation dataset, we extracted segments from all the videos referenced in the evaluation set of AudioSet. For this validation dataset, we applied only the first step of the procedure we used to build the training set. The second step was not necessary given that no label had more than $2000$ segments. In total $582$ segments were selected to download. However due to unavailability to download some segments, the total set became $532$ segments with $657$ total labels. The resulting evaluation set provides at least $53$ examples for each label. The least represented label is \textit{Thunder} with $53$ examples while the most represented is \textit{Wind} with $173$ examples.  Fig.~\ref{fig:histograma_audioset_validation} shows the resulting distribution of retrieved segments for our validation dataset.

\section{Learning Architecture}
\label{sec:network_architectures}

\begin{figure*}[htb]
    \includegraphics[width=1\textwidth]{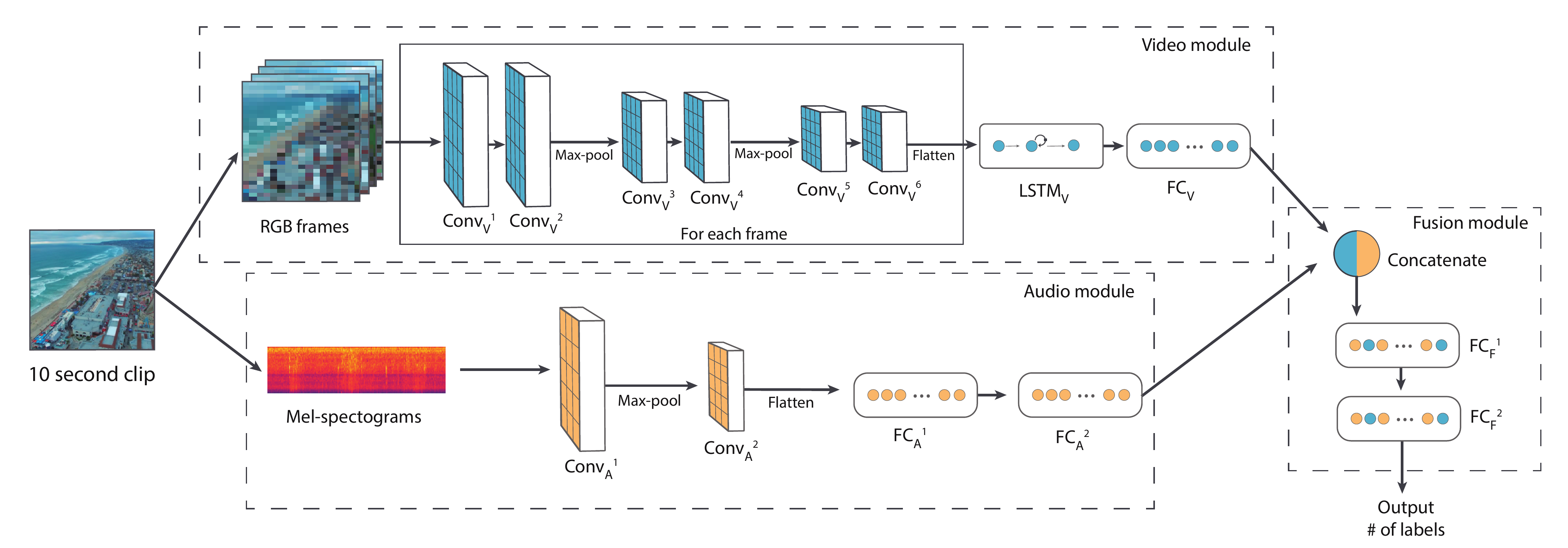}
    \caption{An illustration of our learning architecture, which comprises two components: the bimodal feature extraction net and the fusion net. The bimodal net extracts features from audio and visual signals contained in each training example (clip). The computed feature vectors are concatenated and the resulting vector used as input to a 2-layer fully connected feed-forward network, which does the prediction.}
    \label{fig:learning_architecture}
\end{figure*}

In this section, we describe our proposed neural network architecture to learn a model for the task of predicting scene components within an audiovisual content. Fig.~\ref{fig:learning_architecture} presents an overview of our learning architecture.

As input to our learning architecture, we consider a collection of clips, each one labelled with at least one event class (see Section~\ref{sec:dataset}). Besides, we assume that the audio and visual input signals for each clip used to learn the model are contiguous audio (spectrogram) and video frames, respectively. Each training example is labelled with one or more labels of our select subset of AudioSet.

Our overall bimodal learning architecture comprises three modules (dashed rectangles in Fig.~\ref{fig:learning_architecture}). Two of them are responsible for the bimodal feature extraction, that is, for extracting features from audio and visual signals contained in each training example (clip). The two extracted feature vectors are concatenated and the resulting vector is used as input to the fusion module. We describe the audio and video feature extraction modules in Section~\ref{sec:audio_network} and Section~\ref{sec:video_network}, respectively. In Section~\ref{sec:prediction_module}, we detail the fusion module. Note that, in all three modules, all hidden layers are equipped with ReLU non-linearities~\cite{Nair:2010:RLU:3104322.3104425}.



\subsection{Audio Module}
\label{sec:audio_network}

The architecture used in this network module is somewhat similar to the one employed in~\cite{piczak2015environmental}. The primary architectural difference from the aforementioned implementation is that we used the Mel-spectograms corresponding to an 1-channel input in our network, while the author of~\cite{piczak2015environmental} added their deltas, forming a 2-channel input.



The first convolutional layer ($Conv_A^1$) consists of $80$ filters with $57 \times 6 \times 1$ receptive fields. Then a max-pooling operation is applied with a pooling shape of $4 \times 3$ and stride of $1 \times 3$. In order to avoid overfitting, we apply $50\%$ dropout rate~\cite{Srivastava:2014:DSW:2627435.2670313} after this layer. The second convolutional layer ($Conv_A^2$) consists of $80$ filters with $1 \times 3$ receptive fields. Another max-pooling is applied with a pooling shape of $1 \times 3$ and stride of $1 \times 3$. Batch Normalization~\cite{Ioffe:2015:BNA:3045118.3045167} was applied after each convolutional layer.

The activation volume resulting from $Conv_A^2$ is flattened to a $3600 \times 1$ vector and further fed into a fully connected stage with 2 layers ($FC_A^1$ and $FC_A^2$), the first with $5000$ units, and the second with $1000$ units, to standardize the feature-length dimensionality and feed to the fusion network. A $50\%$ dropout rate is also applied after the first fully connected layer ($\operatorname{FC}_A^1$).

\subsection{Video Module}
\label{sec:video_network}

This module extracts features from a sequence of frames by first applying three convolutional stages to it (see parts labelled $Conv_V^i$, $1\leq i \leq 6$, in Fig.~\ref{fig:learning_architecture}). Each stage is composed of two identical convolutional layers followed by a batch normalization and max-pooling operation with shape of $2\times2$ and stride of $2 \times 2$. All the convolutional filters have $3 \times 3$ receptive fields and pad the output with zeros to keep the same input size in all of the stages. The amount of filters in the first, second and third convolutional stages are $32$, $64$ and $128$, respectively. A 50\% dropout rate is applied after the first convolutional stage.

The goal of this module is to learn a representation for video clips that will enhance the prediction based on audio. For this matter, every convolutional layer is applied in the time dimension as well (i.e., to every frame in the sequence) in order to learn spatio-temporal features. 

The convolutional layer activations are flattened to a $512 \times 1$ shape. This process happens for every frame of the input sequence. The resulting shape is joined with the subsequent frames. The resulting shape is further fed to a Long-Short Term Memory (LSTM) layer~\cite{Hochreiter:1997:LSM:1246443.1246450} with 256 cells to learn the long-term temporal structure of frames. Lastly, this is followed by a fully connected layer ($FC_V$) with $1000$ units. A $50\%$ dropout rate is applied after the first convolutional stage and after the LSTM layer ($\operatorname{LSTM}_V$).

\subsection{Fusion Module}
\label{sec:prediction_module}

The fusion network is modelled to do the prediction task. Thus we concatenate the two vectors resulting from the Audio and Video modules. This concatenation leaves us with a $2000$-dimensional feature vector, which is used as input to the fusion module. The fusion module is a fully connected feed-forward network, which does the prediction. It comprises 2 hidden layers, each of them with $500$ units. A 50\% dropout rate is applied after each hidden layer.

It is possible that more than one label are associated to a single training example in our selected subset of AudioSet (see Section~\ref{sec:dataset}), which results in a classification setting that is both multi-class and multi-label. Hence, the output layer of the fusion module comprises 7 units, one for each label in our dataset. We map the predictions to the labels using a sigmoid output layer. The reason for this choice is that, for our multi-label classification problem, it does not make sense to use softmax in the output layer, since we wanted to produce an independent output for each label. For this reason, we decided to use sigmoid non-linearity in the output layer, since for this function the predicted outputs for each label is in the continuous range [0, 1], where a value near one means the presence of the corresponding event label, and a value near zero means its absence.

We employ a multi-label training loss in the fusion module. Precisely, we use the binary cross entropy loss function, which is adequate for such classification setting.


\section{Experiments and Results}
\label{sec:experiments_and_results}

This Section presents the experiments we performed in order to validate our learning architecture.\footnote{Source code for assembling the dataset and for training the networks can be downloaded from https://github.com/MLRG-CEFET-RJ/bimodal\_audioset} Section \ref{sec:data_preparation} describes the data preprocessing activities. Section~\ref{sec:evaluation_metrics} presents the metrics we used to evaluate our models. Finally, Section~\ref{sec:results_and_analysis} describes the experiments we conduct to validate our learning architecture, along with a corresponding analysis of results.

\subsection{Data preparation}
\label{sec:data_preparation}

In Section~\ref{sec:dataset}, we describe the procedure we followed to build our training and validation datasets. The preparation of segments in these datasets was performed as follows:

\begin{itemize}
    \item \textbf{Video}: In AudioSet, every segment has a fixed sized of 10 seconds. To reduce the problem space, we subsampled every video-clip down to $32$ frames. We also downsized each frame to $32 \times 32$ pixels and kept RGB values, leaving us with $32 \times 32 \times 32 \times 3$ tensor for the video signal of each training example. After that, we normalized the value of each pixel to keep it within the range $[0, 1]$.

    \item \textbf{Audio}: The task of learning temporal information from sound events can be accomplished by feeding the raw signals directly into a neural network, as shown in~\cite{dai2017very}. But as have been verified in~\cite{piczak2015environmental}, some raw audio data can be transformed into lesser complex acoustic characteristics that represent the sound, such as Log-scaled mel-spectrograms. Thus we employ the same technique presented by that author to transform the audio signal of each training example. Firstly, all audio clips were resampled at 22050 Hz and min-max normalized between $-1$ and $1$. Log-scaled Mel spectrograms were extracted from all recordings with window size of $1024$, hop length of $512$ and $60$ mel-bands.  Different from the implementation in~\cite{piczak2015environmental}, we did not split the spectrograms into frames. 
    
    The reason for this design decision is to maintain the same temporal scale for the both inputs in our bimodal architecture. This decision enables the network to learn on the whole 10s clips along with video representation.
\end{itemize}

\subsection{Evaluation Metrics}
\label{sec:evaluation_metrics}

To measure accuracy of the trained models, we utilized several evaluation metrics: Micro-averaged F1-score, Exact Match Ratio, and Hamming Loss. Bellow, we briefly describe each one of them in this section. For the equations presented in this section, consider that $|L|$ is the total number of labels, and $|D|$ is the number of examples in the validation set.

As the datasets utilized in this work are unbalanced we need performance metrics that are independent of the class distribution. Thus we opted to estimate the Micro-averaged F1-score \cite{yang1999evaluation} (Micro-F1) which can be seen as the weighted average of F1 scores over all the labels. Following \cite{tang2009large}, the Micro-F1 can be defined as presented in Eq.~\ref{eq:f1}. In this equation,  $x_i$ is the predicted value for a given example, and $y_i$ is the corresponding ground truth.

\begin{equation}
\operatorname{Micro-F1} = \frac{2\sum_{l=1}^{|L|}\sum_{i=1}^{|D|} x_i^l y_i^l}{\sum_{l=1}^{|L|}\sum_{i=1}^{|D|} x_i^l + \sum_{l=1}^{|L|}\sum_{i=1}^{|D|}y_i^l}
\label{eq:f1}
\end{equation}







Exact Match Ratio ($\operatorname{MR}$)~\cite{kazawa2005maximal} considers one instance as correct if and only if all associated labels are correctly predicted. Exact Match Ratio formula is given by Eq. \ref{eq:mr}, where $|D|$ is defined as above, $y_i$ is the ground truth for the $i$-th example, $x_i$ is the prediction, and $I$ is the indicator function (equals $1$ if the statement $x_i = y_i$ is true, and equals $0$ otherwise).

\begin{equation}
    \operatorname{MR}(x_i, y_i) = \frac{1}{|D|} \sum_{i=1}^{|D|} I [x_i = y_i]
    \label{eq:mr}    
\end{equation}

A disadvantage of $\operatorname{MR}$ criterion is that it does not take partial matches into account. In order to account for partially correctness we also employ Hamming Loss, that represents how many times on average, a label is incorrectly predicted \cite{tsoumakas2007random}. The Hamming Loss formula is presented in Eq.~\ref{eq:hamming}, where $|D|$ and $|L|$ are defined as above, $x_i$ is the predicted value for a given example, and $y_i$ is the corresponding ground truth. The exclusive disjunction ($\operatorname{xor}$ operation) is used to compute the symmetric difference of the values for each label. It returns $0$ if the label is equal in both $x_i$ and $y_i$ and $1$ otherwise.

\begin{equation}
    \operatorname{HammingLoss}(x_i, y_i) = \frac{1}{|D|} \sum_{i=1}^{|D|} \frac{\operatorname{xor}(x_i, y_i)}{|L|}
    \label{eq:hamming}
\end{equation}


\subsection{Results and Analysis}
\label{sec:results_and_analysis}

Network optimization hyper-parameters for all modalities were similar to the ones used in~\cite{piczak2015environmental}. We optimized our models using Stochastic Gradient Descent with Nesterov momentum of $0.9$ and learning rate of $0.01$. Both Video Module and Audio Module were trained for $40$ epochs with mini-batch implementation, with batch size of $32$. We implemented our learning architecture using Keras~\cite{chollet2015keras} with TensorFlow~\cite{abadi2016tensorflow} as backend.

In order to validate our proposed learning architecture (see Section~\ref{sec:network_architectures}), we first investigate three different experimental settings: we train models using audio signal only, video signal only, and both signals (i.e., bimodal).

The first and second experimental settings correspond to training the Audio and Video modules, respectively, as separated networks. Results for these experimental settings are presented in the first three lines of Table~\ref{tab:comparsion}. We call the network trained only on audio features as \textit{Audio Only}, the network trained only on video as \textit{Video Only} and the bimodal network architecture as \textit{Bimodal}.

In order to keep the output of our separate networks analogous to the Bimodal network output, a 50\% dropout was also applied after the last hidden layer of our separate modalities (respectively after $\operatorname{FC}_V$ for Video Only and after $\operatorname{FC}_A^2$ for Audio Only). After the dropout we map the predictions to the labels using a sigmoid output layer in the same manner as described in Section~\ref{sec:prediction_module}.

One could argue that our bimodal architecture produces better accuracy only because of the concatenation and addition of two fully connected layers ($FC_F^1$ and $FC_F^2$ of Fig.~\ref{fig:learning_architecture}). In order to verify this assumption, we investigate two other experimental settings. For the first additional setting, we duplicate the $1000$-dimensional feature vector produced by layer $FC_A^2$, which results in two $1000$-dimensional vectors, say, $v_1$ and $v_2$. After that, we provide vectors $v_1$ and $v_2$ as input to the Fusion Module. For the second additional setting, we repeat the procedure for the output of layer $FC_V$. Results for these experimental settings are presented in the forth and fifth lines of Table~\ref{tab:comparsion}. 
We call `Fusioned Audio' the audio network trained together with the fusion module; and, correspondingly, for the video network, we use `Fusioned Video'.

\begin{table}[htb]
\centering
\caption{Evaluation results for unimodal and bimodal settings.}
\label{tab:comparsion}
\begin{tabular}{l||c|c|c}
\textbf{Model type} & \textbf{Micro-F1} & \textbf{Hamming Loss}  & \textbf{MR}  \\[0.1cm]
\hline
Audio Only & 0.496756 & 0.145811 & 0.33\% \\[0.1cm]
Video Only & 0.509954 & 0.174544 & 0.31\% \\[0.1cm]
Bimodal & 0.639498 & 0.123523 & 0.48\% \\[0.1cm]
\hline
Fusioned Audio & 0.441629 & 0.165682 & 0.32\% \\[0.1cm]
Fusioned Video & 0.466116 & 0.178303 & 0.29\% \\[0.1cm]
\hline
Audio Only + Video Only & 0.593649 & 0.168367 & 0.30\% \\
\end{tabular}
\end{table}

\begin{figure}[htb]
\centering
\includegraphics[height=0.23\textheight]{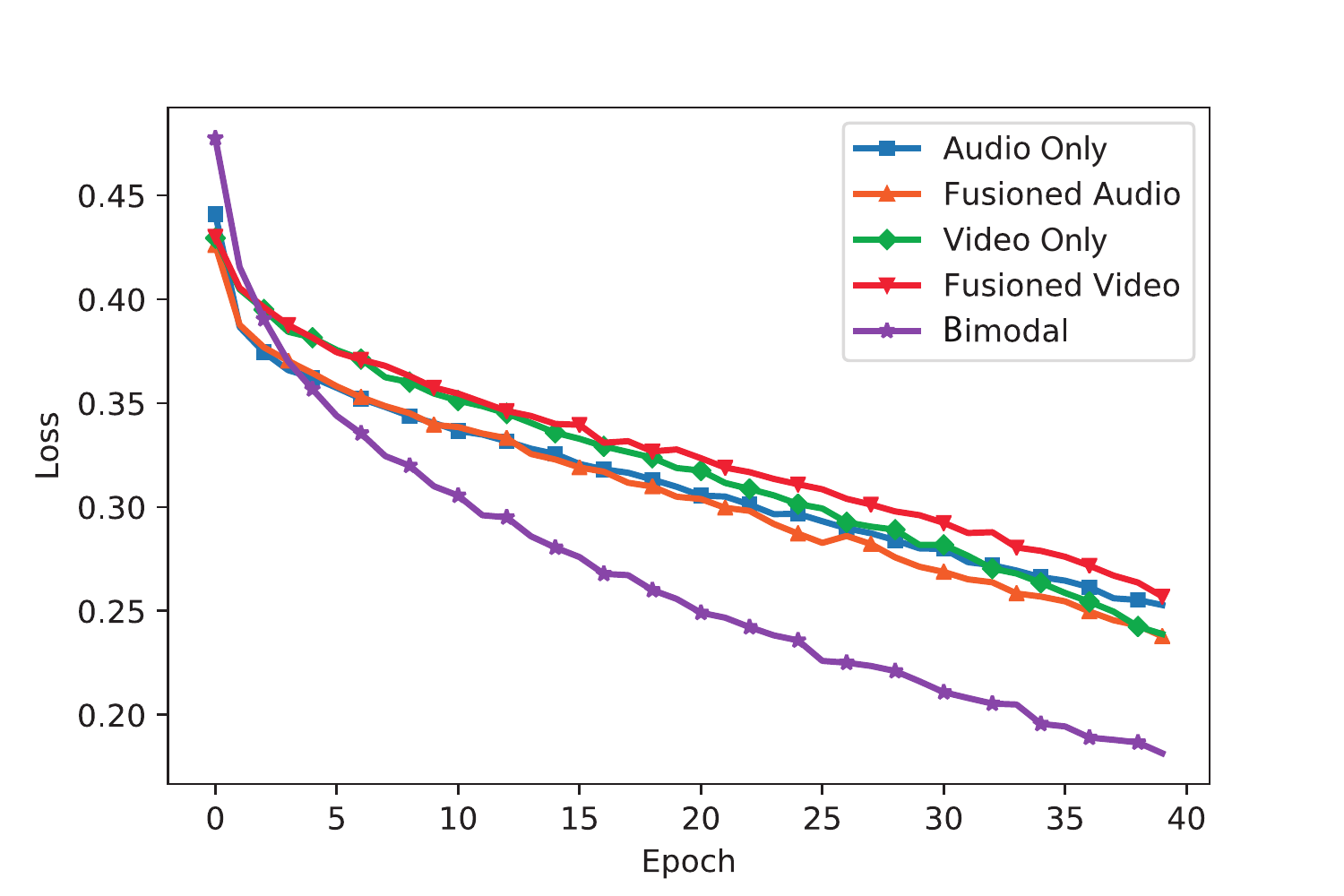}
\caption{Learning curves for unimodal and bimodal settings. Our bimodal learning architecture shows a faster convergence than unimodal networks.}
\label{fig:loss}
\end{figure}

Fig.~\ref{fig:loss} shows the learning curves for all settings (we stopped at $40$ training epochs because models were starting to overfit after this limit). It can be seen that training on our bimodal architecture results in faster convergence.

In a last experiment, we wanted to verify whether the separate modalities with their predictions are combined produce a better accuracy than the bimodal network. For this experimental setting we extracted the prediction vectors for audio and video networks and applied a element-wise sum for every prediction vector. We call such setting \textit{Audio Only + Video Only}. The results for this experimental setting is presented in the sixth line of Table~\ref{tab:comparsion}. 

To assess the results per label, we also evaluated the amount of True Positives (TP), True Negatives (TN), False Positives (FP) and False Negatives (FN) for each label of the main network modalities. Audio only is presented on Table~\ref{tab:audio}, Video only is presented on Table~\ref{tab:video} and the bimodal architecture results is presented on Table~\ref{tab:bimodal}.
We can observe that the label \emph{Fire} was better recognized by the video network, while \emph{Rain} was better recognized by the audio network. Our bimodal network surpasses the predictions of the individual networks for both of these labels. 

We can also observe that, for the label ``Ocean'', the amount of FP were considerably higher in the Video Only network than in the Audio Only network, while the TP rate for the same label was also higher on the Video Only network than on the Audio Only network. On the other hand, the Bimodal network has a much lower ratio between the FP rate and TP high. That indicates that the Bimodal network can correctly predict more labels and thus outperform both separate networks.

\begin{table}[!h]
\centering
\caption{Counts for the audio only setting.}
\label{tab:audio}
\begin{tabular}{l||cccc}
\textbf{Label} & \textbf{TP}  & \textbf{FP}  & \textbf{TN}   & \textbf{FN}  \\
\hline
Wind & 80 & 47 & 312 & 93 \\
Thunder & 15 & 2 & 477 & 38 \\
Rain & 44 & 24 & 416 & 48 \\
Ocean & 17 & 35 & 435 & 45 \\
Fire & 14 & 17 & 455 & 46 \\
Explosion & 18 & 20 & 457 & 37 \\
Gunshot, gunfire & 80 & 9 & 361 & 82 \\
\hline
Total & 268 & 154 & 2913 & 389 
\end{tabular}
\end{table}

\begin{table}[!h]
\centering
\caption{Counts for the video only setting.}
\label{tab:video}
\begin{tabular}{l||cccc}
\textbf{Label} & \textbf{TP}  & \textbf{FP}  & \textbf{TN}   & \textbf{FN}  \\
\hline
Wind & 112 & 80 & 279  & 61 \\
Thunder & 12 &   16 & 463 &   41 \\
Rain & 36 &  33 & 407 &   56 \\
Ocean & 49 &  112 &   358 &   13 \\
Fire & 28 &  12 & 460 &   32 \\
Explosion & 22 &   30 & 447 &   33 \\
Gunshot, gunfire & 74 &  33 & 337 &   88 \\
\hline
Total & 333 & 316 &   2751 &  324
\end{tabular}
\end{table}

\begin{table}[!h]
\centering
\caption{Counts for the bimodal setting.}
\label{tab:bimodal}
\begin{tabular}{l||cccc}
\textbf{Label} & \textbf{TP}  & \textbf{FP}  & \textbf{TN}   & \textbf{FN}  \\
\hline
Wind & 111 & 53 & 306 & 62 \\
Thunder & 28 & 14 & 465 & 25 \\
Rain & 70 & 44 & 396 & 22 \\
Ocean & 41 & 49 & 421 & 21 \\
Fire & 42 & 24 & 448 & 18 \\
Explosion & 21 & 14 & 463 & 34 \\
Gunshot, gunfire & 95 & 13 & 357 & 67 \\
\hline
Total & 408 & 211 & 2856 & 249 
\end{tabular}
\end{table}

\begin{figure}[htb]
\centering
\includegraphics[height=0.3\textheight]{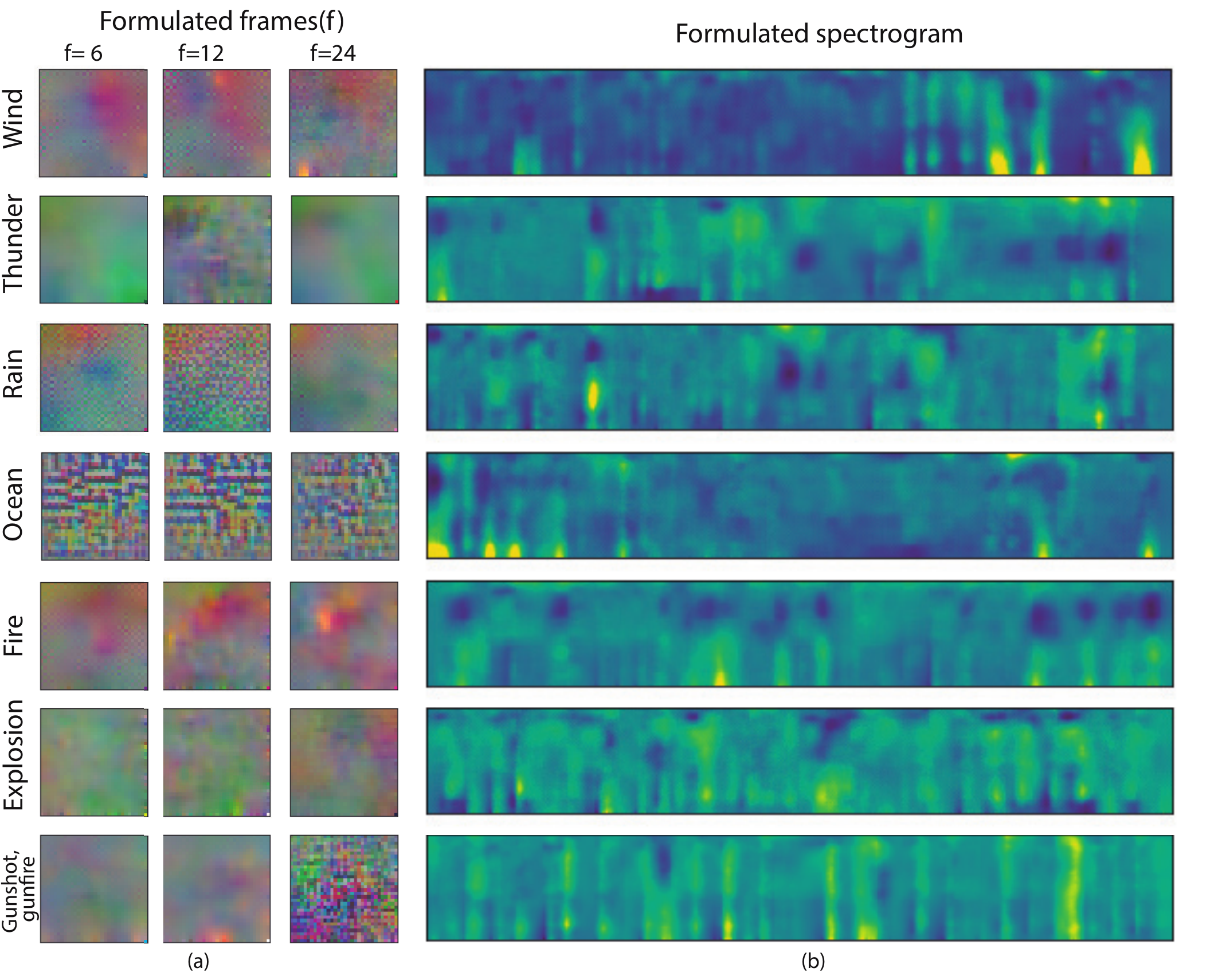}
\caption{Class Model Visualization of Audio and video inputs for each label.}
\label{fig:conv_activation}
\end{figure}

In order to visualize how the network is behaving we formulate images which will best represent each label using Class Model Visualization \cite{simonyan2013deep}. In Fig.~\ref{fig:conv_activation} we present audio and video inputs that try to maximize the output of the network for each label. The left (Fig.~\ref{fig:conv_activation}(a)) subplots represent the inputs for video on frames $6$, $12$ and $24$\footnote{We also invite the reader to consult the accompanying video containing the full animation: https://youtu.be/dTVbsootmiA}. In Fig.~\ref{fig:conv_activation}(b) we present the formulated log-scaled mel-spectogram feature of the recording that maximizes each label. We can see from these visualizations that the network is actually learning the sound events from a relevant time period in the log-scaled mel-spectogram feature. As for video, some labels seem to have distinction between frames as time progresses (e.g., \emph{Fire}, \emph{Gunshot, gunfire}), which indicates that during training the network has captured an understanding of the label presence over time.

\section{Conclusion} 
\label{sec:conclusion}

Manually identifying the timestamps to start the execution of sensory effects in mulsemedia applications is often time consuming and error-prone. In this work, we showed how a bimodal deep learning architecture can be trained in order to solve this task.

Our proposed bimodal architecture comprises three modules, two of them for extracting audio and video features from input videos, and a third one, called fusion module, that combines both results to do the prediction of scene components. In order to validate our architecture, we employed the architecture on a subset of the AudioSet dataset targeting the prediction of scene components. The proposed architecture was validated using six different experimental settings. In all cases, our experiments showed that our bimodal architecture performed better.

We also employed a method to extract a dataset tailored for training predictive models for scene components identification. We extracted a subset derived from the AudioSet dataset to train our models. The main challenge of this first version is the lack of relationship between video and audio in some Youtube clips. An example of this is the existence of clips labelled as \emph{Rain} that encompasses video occurrences from inside houses, games, real life scenarios and static images with only sound of rain effects.  
Another challenge was the training of our networks with multi-labeled data, which means that maybe multiple labels co-exist in the same clip. It is known that the multi-label training method has the potential to find correlations among labels \cite{zhang2014review}. If a real correlation between the labels is present and desired it can help the learning process. On the other hand, if no correlation is desired it could endanger the learning process.





Given the challenges described above, a venue for future work is to apply data cleaning and data augmentation techniques to extend and improve the quality of the dataset presented in this work. As another future work, other learning techniques, such as binary relevance \cite{zhang2014review}, might be explored to seek improvements in the learning process.

We also consider to incorporate deep CCA \cite{andrew2013deep} in the Fusion Module of our learning architecture, in order to take into account correlations between audio and video modalities.

Finally we also plan to publish a scene recognition web service built upon the bimodal learning architecture proposed in this paper.


\section*{Acknowledgment}

The authors would like to thank CNPq, CAPES, and FAPERJ for partially funding this research.

\bibliographystyle{IEEEtran}
\bibliography{referencias}
\end{document}